\documentclass[10pt,twocolumn,letterpaper]{article}

\usepackage{cvpr}              

\usepackage{colortbl}
\usepackage{color}
\usepackage{multirow}
\usepackage{pifont}
\usepackage{makecell}
\definecolor{highlight}{RGB}{58,110,165}

\usepackage{ntheorem}

\usepackage{booktabs}      
\usepackage{multirow}      
\usepackage{graphicx}      
\usepackage{makecell}      
\usepackage{array} 
\usepackage{siunitx}
\usepackage{calc}

\usepackage[table]{xcolor} 
\usepackage{nicematrix} 
\newcolumntype{C}[1]{w{c}{#1}}

\definecolor{cvprblue}{rgb}{0.21,0.49,0.74}
\usepackage[pagebackref,breaklinks,colorlinks,citecolor=cvprblue]{hyperref}

\def\paperID{}
\def\confName{CVPR}
\def\confYear{2026}

\title{Local Precise Refinement: A Dual-Gated Mixture-of-Experts for Enhancing Foundation Model Generalization against Spectral Shifts}

\author{%
Xi Chen ~ Maojun Zhang ~ Yu Liu ~ Shen Yan{\footnotemark[2]}\\
\normalsize
\ National University of Defense Technology\\
\normalsize
{\tt\small \{xi\_chen,mjzhang,jasonyuliu,yanshen12\}@nudt.edu.cn}\\
{\tt\small \url{https://nudt-sawlab.github.io/SpectralMoE/}}
}

\begin{document}

\twocolumn[{%
	\renewcommand\twocolumn[1][]{#1}%
	\maketitle%
    \setlength{\abovecaptionskip}{0.1cm}
    \setlength{\belowcaptionskip}{0.1cm}
	\begin{center}
		\centering
        \vspace{-1.0cm}

    \begin{tabular}{c@{\hspace{-0.00\textwidth}}c} 
		\includegraphics[width=0.3\textwidth]{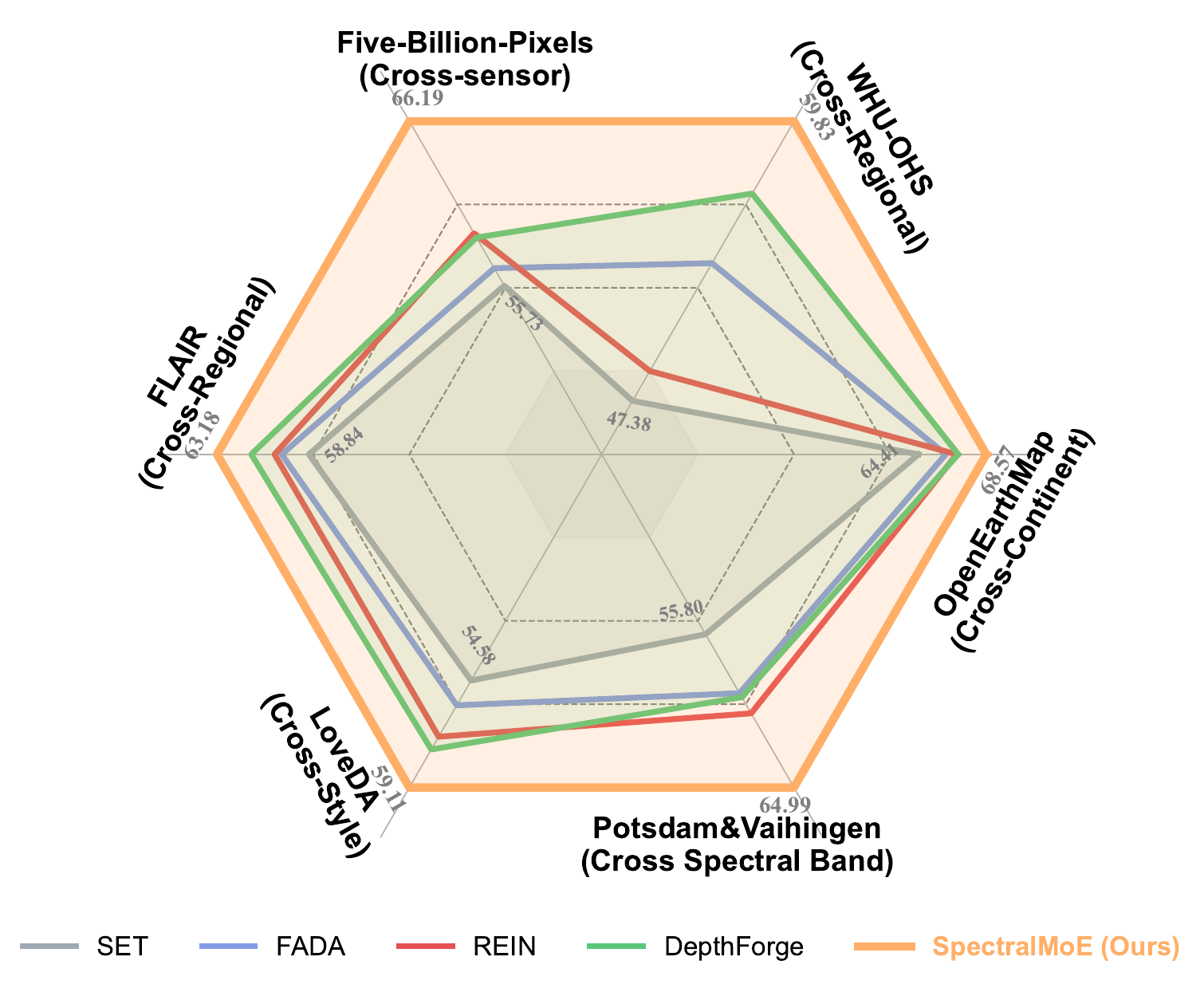}&
		\includegraphics[width=0.68\textwidth]{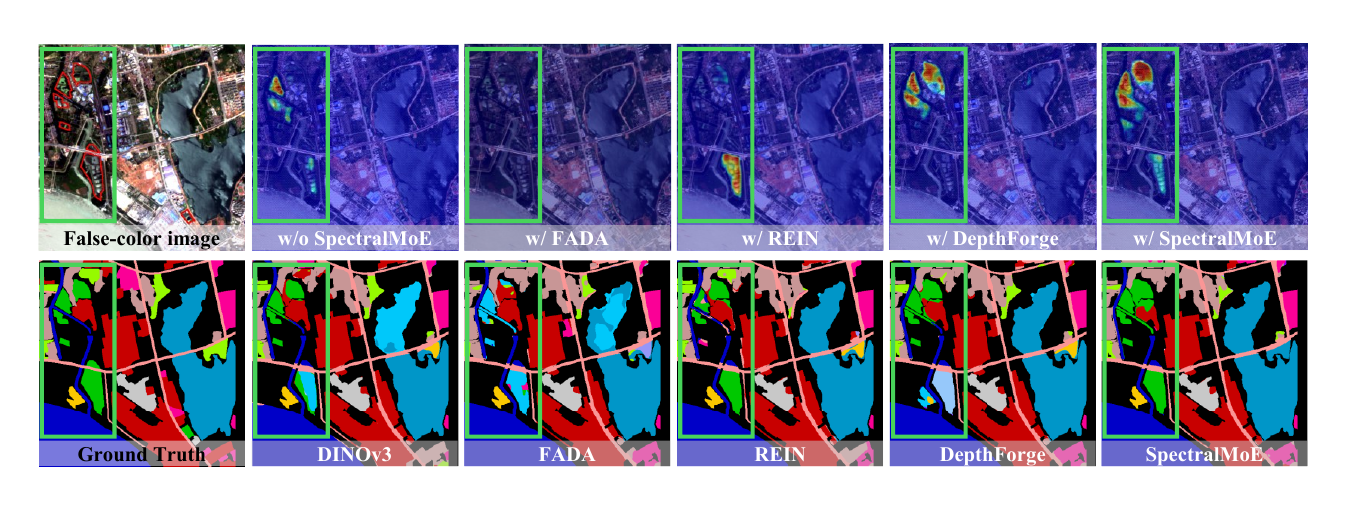}\\
		\footnotesize (a) \textbf{Comprehensive SOTA Performance} &
		\footnotesize (b) \textbf{Fine-Grained Local Adjustment and Superior Generalization} \\
	\end{tabular}
    \captionof{figure}{Our \textbf{SpectralMoE} achieves (a) comprehensive SOTA performance across all spectral RS DGSS benchmarks. This superiority stems from our dual-gated MoE, which enables (b) fine-grained, spatially-adaptive adjustments. As shown in the qualitative results (visualized by Grad-CAM, \textbf{top row}), for complex target regions (e.g., the structures highlighted in the green box), our method generates a complete and fine-grained response, in stark contrast to the diffuse activations of global fine-tuning methods. This enhanced local refinement directly translates to qualitatively superior and more robust segmentation results in unseen domains (\textbf{bottom row}).
    }
    \label{fig:figure1} 
    \vspace{-4.0 mm}
    \end{center}     
}]
\renewcommand{\thefootnote}{\fnsymbol{footnote}}
\footnotetext[2]{~Corresponding authors.}

\begin{abstract}

Domain Generalization Semantic Segmentation (DGSS) in spectral remote sensing is severely challenged by spectral shifts across diverse acquisition conditions, which cause significant performance degradation for models deployed in unseen domains. 
While fine-tuning foundation models is a promising direction, existing methods employ global, homogeneous adjustments. This "one-size-fits-all" tuning struggles with the spatial heterogeneity of land cover, causing semantic confusion.
We argue that the key to robust DGSS lies not in a single global adaptation, but in performing fine-grained, spatially-adaptive refinement of a foundation model's features.
To achieve this, we propose SpectralMoE, a novel fine-tuning framework for DGSS. It operationalizes this principle by utilizing a Mixture-of-Experts (MoE) architecture to perform \textbf{local precise refinement} on the foundation model's features, incorporating depth features estimated from selected RGB bands of the spectral remote sensing imagery to guide the fine-tuning process.
Specifically, SpectralMoE employs a dual-gated MoE architecture that independently routes visual and depth features to top-k selected experts for specialized refinement, enabling modality-specific adjustments.
A subsequent cross-attention mechanism then judiciously fuses the refined structural cues into the visual stream, mitigating semantic ambiguities caused by spectral variations.
Extensive experiments show that SpectralMoE sets a new state-of-the-art on multiple DGSS benchmarks across hyperspectral, multispectral, and RGB remote sensing imagery.
\vspace{-4mm}
\end{abstract}    
\section{Introduction}
\label{sec:intro}

\begin{figure*}[t]
    \centering
	\includegraphics[ width=1.0\linewidth]{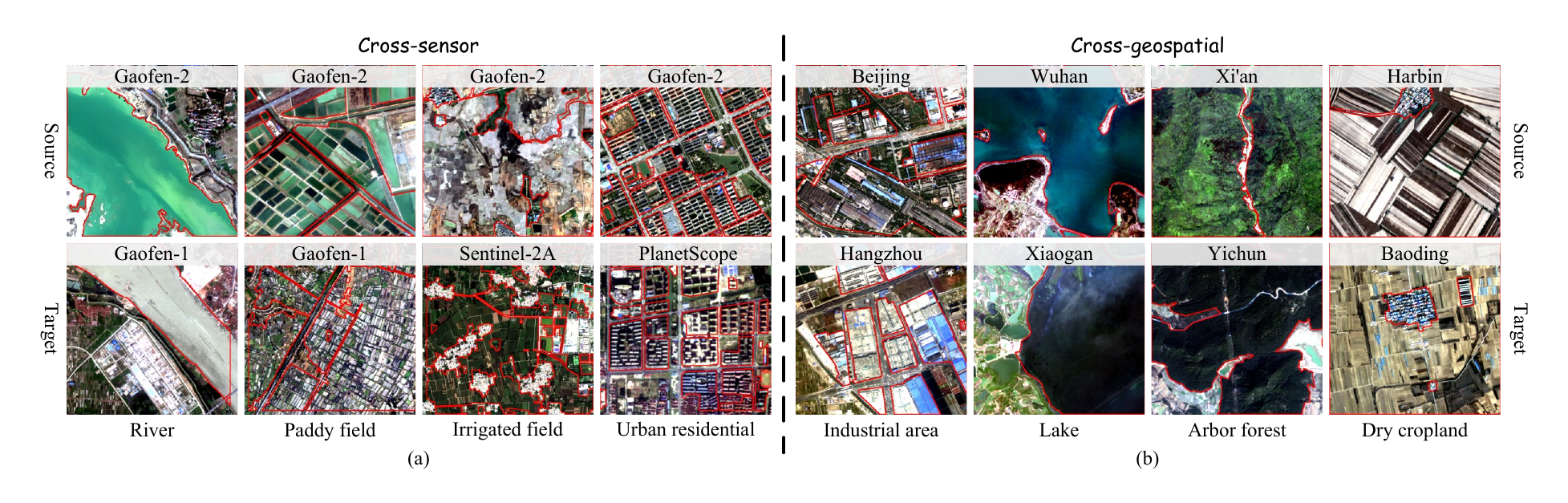}
    \vspace{-25pt}
	\caption{\textbf{Spectral shift in spectral RS imagery.} Variations in sensor characteristics and geospatial conditions can lead to significant divergence in the spectral signatures of land cover features belonging to the same class.}
    \label{fig:introduction}
    \vspace{-4mm}
\end{figure*}

Spectral Remote Sensing (RS) imagery~\cite{10490262}, encompassing hyperspectral, multispectral, and natural RGB data, provides rich information for earth observation. 
The semantic segmentation of these images aims to classify the land cover type of each pixel. 
However, significant spectral shifts~\cite{chen2024spectral}, caused by vast differences in sensor types, illumination, seasons, and geographical environments, make it exceptionally challenging to maintain high segmentation accuracy in unseen scenes (as illustrated in Fig.~\ref{fig:introduction}). 
The task of Domain Generalization Semantic Segmentation (DGSS) for spectral RS is designed to enhance a model's cross-scene generalization ability to unseen geographical regions or imaging conditions~\cite{luo2025domain}, without requiring access to this target data during training. 
This provides a robust generalization capability crucial for the practical application~\cite{peng2026lodlocv3generalizedaerial} of spectral RS image interpretation.

Previous methods have primarily focused on techniques such as data augmentation~\cite{chattopadhyay2023pasta, lee2022wildnet, peng2021global}, domain-invariant feature learning~\cite{xu2022dirl, zhang2023learning, chen2024spectral}, and meta-learning~\cite{kim2022pin, dou2019domain} to achieve good domain generalization performance. 
However, they are often constrained by smaller-scale backbones, which limits the robustness of their generalization capabilities. 
Recently, the development of Foundation Models (FMs)~\cite{bountos2025fomo, li2024s2mae, zhang2025skysense,simeoni2025dinov3, oquab2023dinov2, radford2021learning} has marked a significant breakthrough. 
Pre-trained on either large-scale multi-modal RS data or hundreds of millions of natural images, both RS Foundation Models (RSFMs)~\cite{xiong2024neural,wang2025towards, tseng2025galileo}, such as \textbf{DOFA}~\cite{xiong2024neural}, and Vision Foundation Models (VFMs)~\cite{simeoni2025dinov3, kirillov2023segment, fang2023eva}, like \textbf{DINOv3}~\cite{simeoni2025dinov3}, have demonstrated powerful feature extraction and generalization potential.

Inspired by this, the prevailing paradigm has shifted towards adapting these powerful FMs for downstream segmentation tasks using fine-tuning strategies~\cite{yi2024learning, bi2024learning, wei2024stronger, chen2025stronger}. Nevertheless, existing fine-tuning methods, such as REIN~\cite{wei2024stronger}, typically employ a global and homogeneous strategy for feature adjustment. 
This "one-size-fits-all" approach overlooks the inherent spatial heterogeneity of land cover features in spectral RS imagery. 
For instance, classes like "paddy field" and "pond" are not only spectrally similar but also spatially adjacent. 
A global feature enhancement intended for one class might inadvertently interfere with the features of a neighboring, spectrally similar but semantically distinct class, leading to inter-class confusion. 
This inflexibility in performing fine-grained, differential adjustments to local features poses a key bottleneck for tackling complex spectral RS scenes.

To overcome this limitation, we argue for a paradigm shift from homogeneous adjustments to spatially-adaptive conditional computation. 
Inspired by the Mixture-of-Experts (MoE) model~\cite{NEURIPS2022_91edff07, li2022sparse}, we advocate for independently inspecting the feature vector at each spatial location and dispatching it to the most suitable "expert" network to receive a customized feature adjustment. 
Furthermore, to mitigate the inherent spectral ambiguity in spectral RS images, we introduce spatial structure information as a robust prior. 
Specifically, we leverage a Depth Foundation Model (DFM)~\cite{Lin_2025_CVPR, NEURIPS2024_26cfdcd8}, e.g., PromptDA~\cite{Lin_2025_CVPR}, to infer implicit depth features from the RGB bands of the spectral RS images. 
We posit that this structural information, which encodes the height, contours, and spatial relationships of land cover features, is more robust across different scenes, lighting conditions, and seasons compared to spectral features, making it a crucial element for enhancing model generalization.

To this end, we propose \textbf{SpectralMoE}, a novel fine-tuning framework specifically designed for DGSS in spectral RS. 
It employs a dual-gated MoE mechanism to achieve fine-grained, localized adjustments for both visual and depth-derived structural features. 
Concurrently, it utilizes a cross-attention mechanism to adaptively fuse the spatial structure information from the depth features into the visual features. 
This fusion significantly reduces the semantic ambiguity caused by spectral similarity, ultimately achieving high-precision, cross-domain land cover segmentation for spectral RS imagery, as illustrated in Fig.~\ref{fig:figure1}(b).

Our main contributions are summarized as follows:
\begin{itemize}
    \item We propose SpectralMoE, a novel fine-tuning framework that achieves state-of-the-art DGSS performance on diverse spectral RS benchmarks (hyperspectral, multispectral, and RGB).

    \item We design a dual-gated MoE for local precise refinement. It independently routes visual and depth features, overcoming the inter-class confusion caused by global, homogeneous fine-tuning methods.
    
    \item We introduce robust structural priors (depth estimated from RGB bands) and a cross-attention mechanism to fuse them, effectively mitigating semantic ambiguity caused by spectral similarity.
\end{itemize}

\begin{figure*}[httb]
    \centering
	\includegraphics[ width=1.0\linewidth]{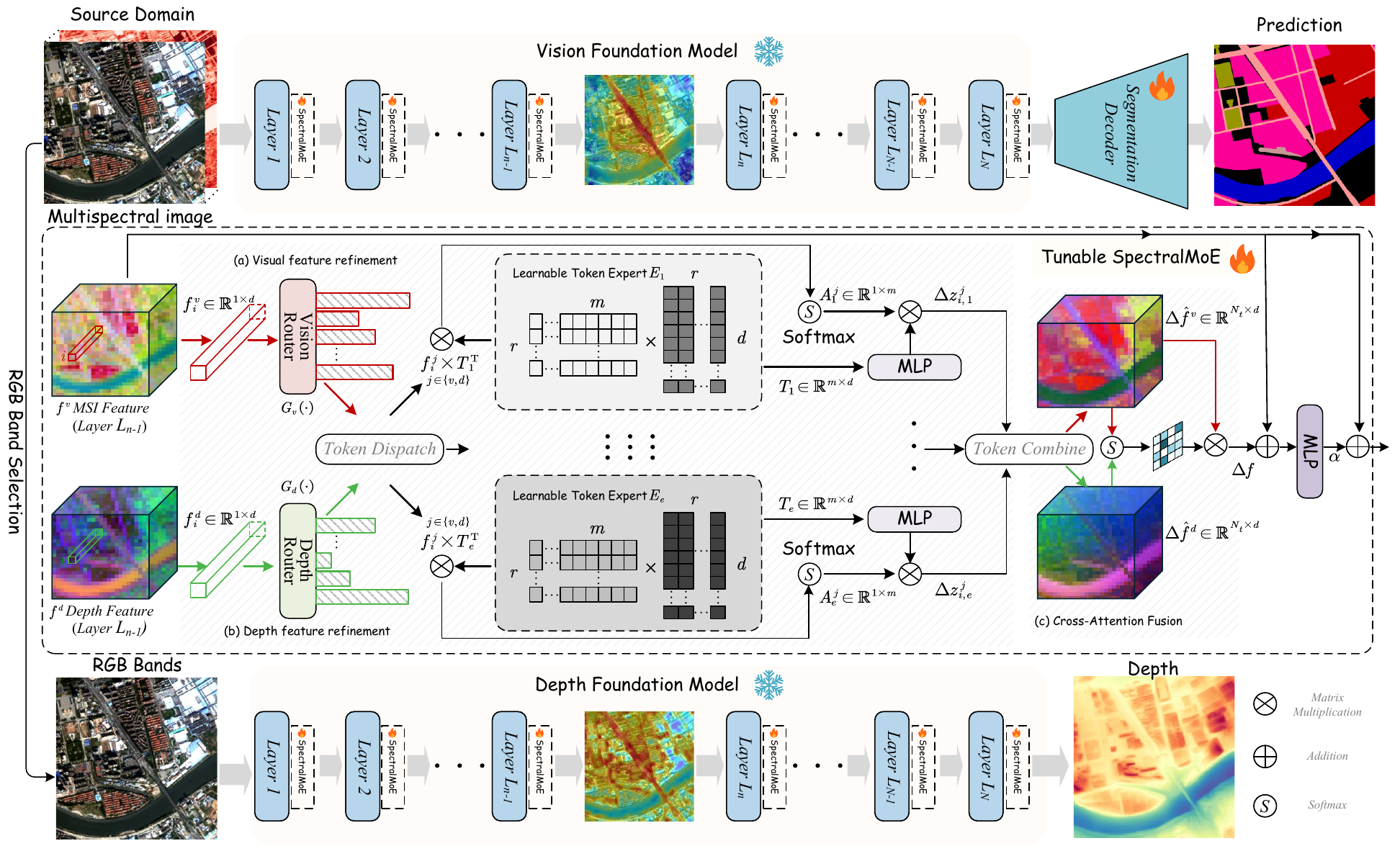}
    \vspace{-15pt}
	\caption{\textbf{Overview of the proposed SpectralMoE framework.} SpectralMoE is inserted as a lightweight plugin into each layer of frozen VFMs and DFMs. At its core is a \textbf{dual-gated MoE} mechanism. A dual-gated network independently routes visual and depth feature tokens to specialized experts, enabling fine-grained, spatially-adaptive adjustments that overcome the limitations of global, homogeneous methods. Following this expert-based refinement, a \textbf{Cross-Attention Fusion Module} adaptively injects the robust spatial structural information from the adjusted depth features into the visual features. This fusion process effectively mitigates semantic ambiguity caused by spectral shifts, significantly enhancing the model's cross-domain generalization capability.}
    \label{fig:method}
    \vspace{-3.5mm}
\end{figure*}
\section{Related Works}
\label{sec:related_works}
\subsection{Domain Generalization Semantic Segmentation for Spectral Remote Sensing}
Domain Generalization Semantic Segmentation (DGSS) in spectral Remote Sensing (RS) aims to train a model on source domains that can generalize effectively to unseen target domains. 
Prior work has explored various strategies, such as data augmentation~\cite{chattopadhyay2023pasta, lee2022wildnet, peng2021global}, learning domain-invariant features~\cite{xu2022dirl, zhang2023learning, chen2024spectral}, and meta-learning~\cite{kim2022pin, dou2019domain}. 
However, the challenge is significantly amplified in spectral RS due to drastic spectral shifts arising from diverse imaging sensors, geographical environments, and lighting conditions~\cite{guo2024skysense}. 
Conventional domain generalization methods~\cite{bi2024learning, ding2023hgformer, choi2021robustnet, 11397394, 11164534}, which typically rely on small- to medium-scale backbones, possess limited representation capacity to model such complex domain variations. 
Consequently, they often fail to generalize robustly to real-world scenes, making the leveraging of powerful Foundation Models to overcome the spectral shift problem a critical and promising research direction.

\subsection{Remote Sensing and Vision Foundation Models}

Foundation Models (FMs)~\cite{bountos2025fomo, li2024s2mae, zhang2025skysense,simeoni2025dinov3, oquab2023dinov2, radford2021learning}, such as Vision Foundation Models (VFMs)~\cite{simeoni2025dinov3, kirillov2023segment, fang2023eva} like DINOv3~\cite{simeoni2025dinov3} and Remote Sensing FMs (RSFMs)~\cite{xiong2024neural, wang2025towards, tseng2025galileo, 10376630} like DOFA~\cite{xiong2024neural}, are pre-trained on massive datasets and produce highly generalizable features. 
Fine-tuning~\cite{yi2024learning, bi2024learning, wei2024stronger, chen2025stronger} is the standard approach to adapt these models for downstream tasks like DGSS. However, existing fine-tuning methods~\cite{hu2022lora, chen2022adaptformer, jia2022visual}, exemplified by methods developed for natural imagery, often fall short when applied to spectral RS. 
The more severe spectral shifts and complex spatial heterogeneity inherent in RS data demand a more powerful and spatially-adaptive feature refinement mechanism. 
Consequently, designing a fine-tuning framework that effectively adapts both VFMs and RSFMs to the unique challenges of domain generalization in spectral RS remains a critical open research direction.


\section{Methodology}
\label{sec:method}
Given a source domain $\mathcal{S} = \left\{ (x_i, y_i) \right\}_{i=1}^{N_s}$ and a target domain $\mathcal{D} = \left\{ x_i \right\}_{i=1}^{N_t}$ with a different distribution, our objective is to train a model on $\mathcal{S}$ that generalizes effectively to $\mathcal{D}$ for spectral RS semantic segmentation.
This task is driven by severe spectral shifts and complex spatial heterogeneity~\cite{TONG2020111322, Li_2021_CVPR, Wang_2021_ICCV}. While fine-tuning FMs is a prevailing strategy, existing methods typically rely on global, homogeneous adjustments. This "one-size-fits-all" approach fails to address local-level challenges, leading to severe inter-class confusion among spectrally similar land cover types.

To overcome this bottleneck, we propose \textbf{SpectralMoE}, a fine-tuning framework that pivots from global tuning to spatially-adaptive, conditional computation. Our approach mainly introduces two innovations: it employs a MoE architecture to perform local precise refinement (addressing spatial heterogeneity) and leverages robust, depth-derived structural priors (mitigating spectral ambiguity). 
Specifically, we insert the lightweight SpectralMoE module as a plugin into each layer of a frozen, pre-trained VFM and a DFM. This module uses a dual-gated MoE for fine-grained, localized adjustments and a cross-attention mechanism to fuse the depth information, enabling fine-grained feature manipulation for highly accurate DGSS. An overview of SpectralMoE is presented in Fig.~\ref{fig:method}.

\subsection{Fine-Grained Feature Adjustment via MoE}
To overcome the limitation of global, homogeneous adjustments in existing methods, we introduce a MoE mechanism. 
Instead of learning a single set of shared adapter tokens for the entire feature map, we instantiate $N_e$ parallel expert networks within each layer of SpectralMoE. 
We then employ a dual-gated network to route visual and depth features token by token, facilitating fine-grained feature modulation.

\noindent \textbf{Expert Parameterization.}
To ensure each expert $E_e$ can perform efficient and targeted feature adjustments, its core parameters are designed as a set of adaptive tokens $T_e \in \mathbb{R}^{m \times d}$. To maximize parameter efficiency, we parameterize these tokens using Low-Rank Decomposition. Specifically, for each expert $E_e$, we learn two low-dimensional matrices: $A_e \in \mathbb{R}^{m \times r}$ and $B_e \in \mathbb{R}^{r \times d}$, where the rank $r \ll d$. The adaptive tokens are then generated by their product: $T_e = A_e \cdot B_e$. This design significantly reduces the number of learnable parameters while preserving the model's expressive capacity.

\noindent \textbf{Dual-Gated Network.}
Considering the inherent representational differences between the visual and depth modalities, using a single shared routing network could lead to mutual interference in routing decisions due to conflicting modal characteristics. To enable differential routing for visual features $f^v$ and depth features $f^d$, we devise a parallel dual-gated network. This network comprises two independent, learnable gating weight matrices, $W_{\text{gate}}^{v} \in \mathbb{R}^{d \times N_e}$ and $W_{\text{gate}}^{d} \in \mathbb{R}^{d \times N_e}$, and two independent noise weight matrices, $W_{\text{noise}}^{v} \in \mathbb{R}^{d \times N_e}$ and $W_{\text{noise}}^{d} \in \mathbb{R}^{d \times N_e}$, corresponding to the visual and depth gating networks, respectively.

The core principle is to establish independent routing pathways for visual and depth features, thereby enabling differentiated and interference-free expert assignment for different modalities. This fine-grained routing mechanism aims to precisely dispatch the feature vector of each modality at any given spatial location to its most suitable set of experts for highly specialized local feature adjustments.

Specifically, for a feature vector $f_i^j$ from the $i$-th spatial position and modality $j \in \{v, d\}$, we compute a vector of routing logits, denoted as $h_i^j$, using a distance-based noisy gating function. The $e$-th component of this vector, $(h_i^j)_e$, is calculated as:
\begin{equation}
(h_{i}^{j})_e = -\| f_{i}^{j} - w_{\text{gate},e}^{j} \|_p + \epsilon_e \cdot \mathrm{Softplus}((f_{i}^{j})^\top w_{\text{noise},e}^{j})
\label{eq:1}
\end{equation}
where $w_{\text{gate},e}^{j}$ and $w_{\text{noise},e}^{j}$ are the $e$-th columns of the gating matrix $W_{\text{gate}}^{j}$ and noise matrix $W_{\text{noise}}^{j}$ for modality $j$, representing the prototype for the $e$-th expert. The term $\| \cdot \|_p$ denotes the $L_p$-norm, used to measure the distance between the feature and the expert prototype; when $p=1$, this corresponds to Laplacian Gating~\cite{NEURIPS2024_7d62a85e}. $\epsilon_e \sim \mathcal{N}(0,1)$ is noise sampled from a standard normal distribution, and $\mathrm{Softplus}(\cdot) = \log(1+\exp(\cdot))$ is the activation function. By using dedicated weight matrices $W_{\text{gate}}^{j}$ and $W_{\text{noise}}^{j}$ for each modality $j$, we ensure the logit computation is entirely modality-independent.

After obtaining the logit vector $h_i^j$, we identify the top-$k$ experts with the highest scores. Let $\mathcal{I}_i^j = \mathrm{arg\,topk}(h_i^j, k)$ be the index set of these $k$ experts. The final gating value $g_{i,e}^j$ assigned to the $e$-th expert is obtained by applying a Softmax normalization to the scores of these selected experts. This process is defined by the piecewise function:
\begin{equation}
g_{i,e}^{j} = 
\begin{cases} 
\frac{\exp(h_{i,e}^{j})}{\sum_{l \in \mathcal{I}_i^j} \exp(h_{i,l}^{j})} & \text{if } e \in \mathcal{I}_i^j \\
0 & \text{if } e \notin \mathcal{I}_i^j
\end{cases}
\label{eq:2}
\end{equation}
where $k \le N_e$. If an expert is not selected, its gate value is zero, and it does not participate in the subsequent feature processing. Through this mechanism, the dual-gated network adaptively matches the most suitable combination of experts for local features of each modality according to their unique characteristics, thereby achieving fine-grained, efficient, and interference-free multi-modal feature refinement.

\noindent \textbf{Expert-based Feature Adjustment.}
Based on the routing results from the dual-gated network, each feature token $f_i^j$ is processed by its selected top-$k$ experts to compute a fine-grained adjustment. Specifically, for a token $f_i^j$ at spatial position $i$, it is processed by each of the $k$ selected experts $E_e$ (where $e \in \mathcal{I}_i^j$). Within expert $E_e$, its adaptive tokens $T_e \in \mathbb{R}^{m \times d}$ are used to modulate the incoming token $f_i^j$. First, a perceptual map $A_{i,e}^{j}$ is computed between the token and the expert's adaptive tokens:
\begin{equation}
A_{i,e}^{j} = \mathrm{Softmax}\left( \frac{f_{i}^{j} \cdot T_e^{\mathrm{T}}}{\sqrt{d}} \right)
\end{equation}
Next, the adaptive tokens $T_e$ are projected via an MLP and element-wise multiplied with the perceptual map $A_{i,e}^{j}$ to produce a customized intermediate adjustment $\Delta z_{i,e}^{j}$ from this specific expert:
\begin{equation}
\Delta z_{i,e}^{j} = A_{i,e}^{j} \cdot \left( T_e \cdot W_T + b_T \right)
\end{equation}
where $W_T \in \mathbb{R}^{d \times d}$ and $b_T \in \mathbb{R}^{d}$ are shared across all experts, with broadcasting applied to $b_T$ during addition.
Finally, the outputs from the top-$k$ selected experts are aggregated using their corresponding gating weights $g_{i,e}^j$ to form the consolidated feature adjustment $\Delta \hat{f}_i^j$ for the spatial position $i$:
\begin{equation}
\Delta \hat{f}_i^j = \sum_{e \in \mathcal{I}_i^j} g_{i,e}^j \cdot \Delta z_{i,e}^j
\end{equation}
This $\Delta \hat{f}_i^j$ represents the fine-grained adjustment for the token $f_i^j$ of modality $j$.

\begin{table}[!tb]
\centering
\footnotesize
\setlength{\tabcolsep}{3.0pt} 
\sisetup{detect-weight, mode=text}
\renewrobustcmd{\bfseries}{\fontseries{b}\selectfont}

\begin{tabular}{@{} l l S[table-format=2.2] S[table-format=2.2] S[table-format=2.2] S[table-format=2.2] S[table-format=2.2] @{}}
\toprule
\multirow{3}{*}{VFM} & \multirow{3}{*}{Method} & {\multirow{3}{*}{\makecell{Params \\ (M)}}} & \multicolumn{4}{c}{mIoU (\%)} \\ 
\cmidrule(lr){4-7} 
& & & {\makecell{FBP \\ \footnotesize(CS)}} & {\makecell{FBP \\ \footnotesize(CR)}} & {\makecell{FLAIR \\ \footnotesize(CR)}} & {Avg.} \\ 
\midrule

\multirow{6}{*}{\makecell[l]{CLIP \\ \footnotesize(Large)~\cite{radford2021learning}}} 
& Freeze & 0.00 & 46.30 & 48.10 & 56.81 & 50.40 \\
& + SET~\cite{yi2024learning} & 6.13 & 55.00 & 52.62 & 59.88 & 55.83 \\
& + FADA~\cite{bi2024learning} & 11.65 & 55.03 & 51.86 & 60.52 & 55.80 \\
& + REIN~\cite{wei2024stronger} & 2.99 & 56.70 & 53.37 & 60.97 & 57.01 \\
& + DepthForge~\cite{chen2025stronger} & 2.99 & 54.77 & 53.89 & 61.17 & 56.61 \\
\rowcolor{gray!10}
& + SpectralMoE & 5.74 & \bfseries 61.33 & \bfseries 54.13 & \bfseries 61.47 & \bfseries 58.98 \\
\addlinespace

\multirow{6}{*}{\makecell[l]{SAM \\ \footnotesize(Huge)~\cite{kirillov2023segment}}}
& Freeze & 0.00 & 44.98 & 47.14 & 56.65 & 49.59 \\
& + SET~\cite{yi2024learning} & 9.21 & 50.31 & 52.10 & 58.17 & 53.53 \\
& + FADA~\cite{bi2024learning} & 16.59 & 49.74 & 50.04 & 59.38 & 53.05 \\
& + REIN~\cite{wei2024stronger} & 4.51 & 50.93 & 50.27 & 59.36 & 53.52 \\
& + DepthForge~\cite{chen2025stronger} & 4.51 & 48.15 & 52.08 & 59.26 & 53.16 \\
\rowcolor{gray!10}
& + SpectralMoE & 10.29 & \bfseries 59.16 & \bfseries 56.79 & \bfseries 59.41 & \bfseries 58.45 \\
\addlinespace

\multirow{6}{*}{\makecell[l]{EVA02 \\ \footnotesize(Large)~\cite{fang2024eva, fang2023eva}}}
& Freeze & 0.00 & 45.48 & 49.33 & 57.52 & 50.78 \\
& + SET~\cite{yi2024learning} & 6.13 & 46.29 & 51.57 & 50.38 & 49.41 \\
& + FADA~\cite{bi2024learning} & 11.65 & 46.49 & 51.27 & 59.37 & 52.38 \\
& + REIN~\cite{wei2024stronger} & 2.99 & 50.81 & 51.33 & 59.46 & 53.87 \\
& + DepthForge~\cite{chen2025stronger} & 2.99 & 40.90 & 47.60 & 58.81 & 49.10 \\
\rowcolor{gray!10}
& + SpectralMoE & 5.74 & \bfseries 60.20 & \bfseries 56.29 & \bfseries 62.09 & \bfseries 59.53 \\
\addlinespace

\multirow{6}{*}{\makecell[l]{DINOv2 \\ \footnotesize(Large)~\cite{oquab2023dinov2}}}
& Freeze & 0.00 & 53.37 & 52.48 & 58.38 & 54.74 \\
& + SET~\cite{yi2024learning} & 6.13 & 55.73 & 55.61 & 58.84 & 56.73 \\
& + FADA~\cite{bi2024learning} & 11.65 & 56.84 & 55.35 & 60.12 & 57.44 \\
& + REIN~\cite{wei2024stronger} & 2.99 & 59.06 & 55.27 & 60.46 & 58.26 \\
& + DepthForge~\cite{chen2025stronger} & 2.99 & 58.79 & 54.92 & 61.56 & 58.42 \\
\rowcolor{gray!10}
& + SpectralMoE & 5.74 & \bfseries 63.77 & \bfseries 57.77 & \bfseries 62.62 & \bfseries 61.39 \\

\bottomrule
\end{tabular}
\vspace{-2mm}
\caption{
    \textbf{Robustness of SpectralMoE across various VFMs.} Abbreviations: Five-Billion-Pixels (FBP), Cross-Sensor (CS), Cross-Regional (CR). Best results for each VFM are in \textbf{bold}.
}
\label{tab:diffVFM}
\vspace{-4mm}
\end{table}

\subsection{Cross-Attention Guided Fusion with Structural Priors}

The token-level adjustments $\Delta \hat{f}_i^v$ and $\Delta \hat{f}_i^d$ from the dual-gated MoE networks are first aggregated into complete feature maps, denoted as $\Delta \hat{f}^v, \Delta \hat{f}^d \in \mathbb{R}^{N_t \times d}$.

To fuse these maps and inject robust spatial priors from the depth features, we use a cross-attention module. We treat the visual adjustment map $\Delta \hat{f}^v$ as the query, and the entire depth adjustment map $\Delta \hat{f}^d$ as both the key and value. This allows all visual adjustment tokens to query the entire depth adjustment map and aggregate relevant structural information. The fusion is formulated as:
\begin{equation}
\Delta f = \mathrm{softmax}\left( \frac{\Delta \hat{f}^v \cdot (\Delta \hat{f}^d)^{\mathrm{T}}}{\sqrt{d}} \right) \cdot \Delta \hat{f}^d
\end{equation}
The resulting fused feature adjustment map $\Delta f$ is then integrated with the original visual feature map $f^v$. This is done via a residual connection, where the sum $\Delta f + f^v$ is processed by an MLP and modulated by a learnable scalar $\alpha$:
\begin{equation}
f_{\text{out}}^{v} = f^{v} + \alpha \cdot \mathrm{MLP}\left( \Delta f + f^{v} \right)
\end{equation}
This process dynamically enriches the visual features with structural depth priors, helping to mitigate ambiguity in spectrally similar regions.

\begin{figure*}[httb]
    \centering
	\includegraphics[ width=1.0\linewidth]{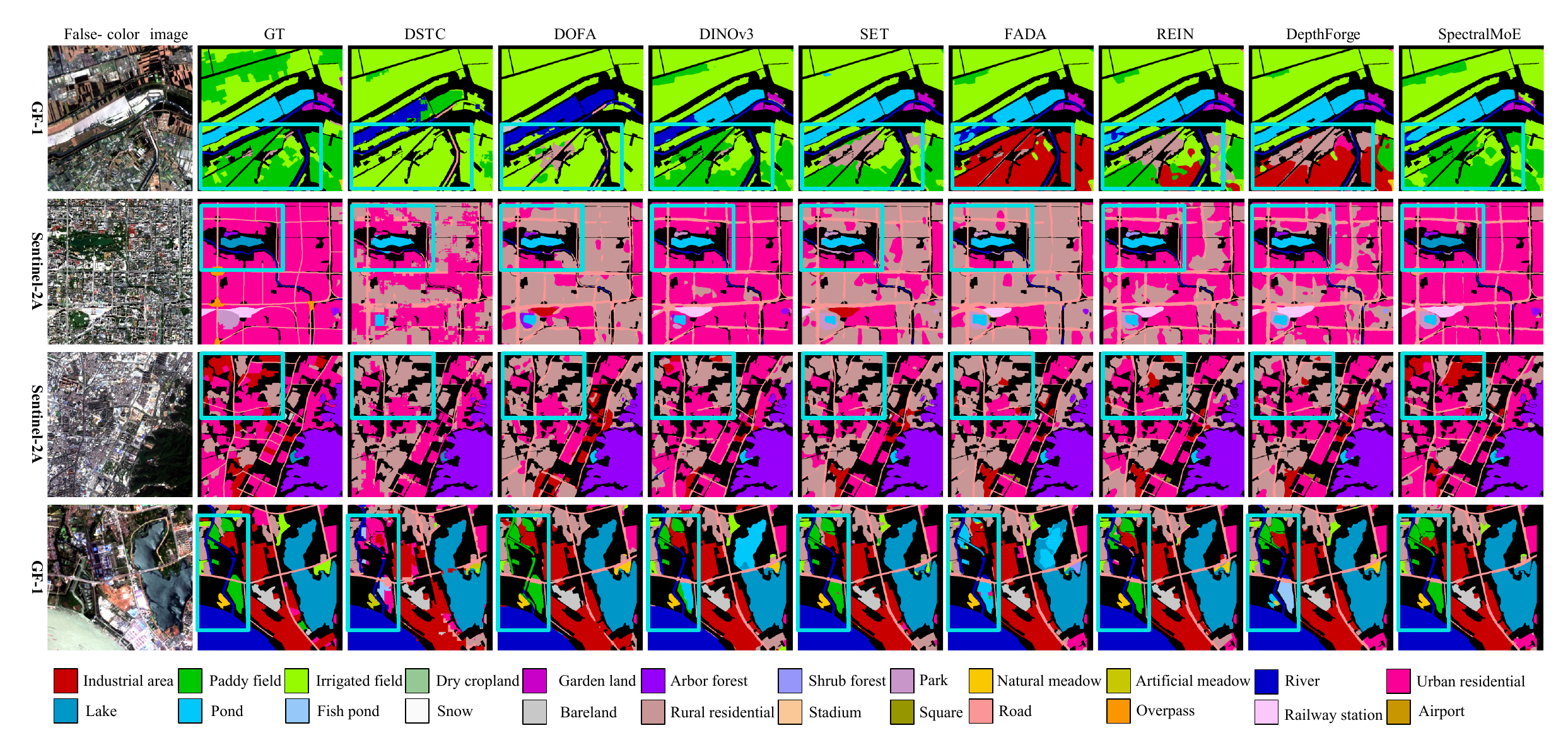}
    \vspace{-20pt}
	\caption{\textbf{Qualitative results for Five-Billion-Pixels (cross-sensor) task.} Visual comparison of segmentation performance for different methods on the Five-Billion-Pixels (cross-sensor) generalization task. From left to right, the columns show the input image, ground truth, and the predictions of DSTC, DOFA, DINOv3, SET, FADA, REIN, DepthForge, and our proposed SpectralMoE. SpectralMoE demonstrates superior generalization capabilities, producing more accurate and refined segmentation maps compared to other SOTA domain generalization methods.}
    \label{fig:GID}
    \vspace{-8mm}
\end{figure*}

\subsection{Optimization Objective}
Given a frozen VFM with parameters $\varPhi_V$ and a DFM with parameters $\varPhi_D$, our goal is to learn the parameters of our SpectralMoE module ($\theta_s$) and a segmentation head ($\theta_h$) by minimizing the following objective:
\begin{equation}
\underset{\theta_s, \theta_h}{\mathrm{arg}\min} \sum_{i=1}^{N_s} \mathcal{L}\left( \mathcal{H}_{\theta_h}\left( \mathcal{F}_{\varPhi_V, \varPhi_D, \theta_s}(x_i) \right), y_i \right)
\end{equation}
where $\mathcal{F}$ represents the forward pass modulated by our SpectralMoE strategy. The total loss $\mathcal{L}$ combines the Mask2Former decoder loss~\cite{cheng2021per} ($\mathcal{L}_{\text{mask}}$) with an auxiliary load-balancing loss ($\mathcal{L}_{\text{load}}$):
\begin{equation}
\mathcal{L} = \mathcal{L}_{\text{mask}} + \lambda \cdot \mathcal{L}_{\text{load}}
\end{equation}
The total load-balancing loss $\mathcal{L}_{\text{load}}$ is the average of the per-modality losses (for visual $f^v$ and depth $f^d$), defined as $\mathcal{L}_{\text{load}} = \frac{1}{2} \sum_{j \in \{v, d\}} \mathcal{L}_{\text{load}}(f^j)$.
This per-modality loss $\mathcal{L}_{\text{load}}(f^j)$ encourages the gating network to distribute the computational load evenly across all experts, preventing expert collapse. We adopt the squared coefficient of variation of expert importance as this loss, expressed as:
\begin{equation}
\mathcal{L}_{\text{load}}(f^j) = \left( \frac{\mathrm{Std}(\mathrm{Imp}(f^j))}{\mathrm{Mean}(\mathrm{Imp}(f^j))} \right)^2
\end{equation}
where, for a given feature $f^j$ (of modality $j$) with $N_t$ tokens, $\mathrm{Imp}(f^j)$ is the set of importances for all experts $\{ \mathrm{Imp}_m^j(f^j) \}_{m=1}^{N_e}$. The importance of a specific expert $m$, denoted as $\mathrm{Imp}_m^j(f^j)$, is calculated by summing its gate values $g_{i,m}^j$ over all $i=1, \dots, N_t$ tokens:
\begin{equation}
\mathrm{Imp}_m^j(f^j) = \sum_{i=1}^{N_t} g_{i,m}^j
\end{equation}
During training, only the parameters of the SpectralMoE module ($\theta_s$) and the segmentation head ($\theta_h$) are updated, keeping the VFM and DFM backbones ($\varPhi_V, \varPhi_D$) frozen.

\begin{table*}[htbp]
\centering
\sisetup{detect-weight, mode=text}
\renewrobustcmd{\bfseries}{\fontseries{b}\selectfont}
\scriptsize
\setlength{\tabcolsep}{4.5pt} 

\begin{tabular}{@{} l *{7}{c} @{}}
\toprule
\multirow{3}{*}{Method} & \multicolumn{1}{c}{\textit{Hyperspectral DGSS Tasks}} & \multicolumn{3}{c}{\textit{Multispectral DGSS Tasks}} & \multicolumn{3}{c}{\textit{RGB DGSS Tasks}} \\
\cmidrule(lr){2-2} \cmidrule(lr){3-5} \cmidrule(lr){6-8}
& \multicolumn{1}{c}{\makecell{WHU-OHS \\ (Cross-Regional)}} & \multicolumn{1}{c}{\makecell{Five-Billion-Pixels \\ (Cross-sensor)}} & \multicolumn{1}{c}{\makecell{Five-Billion-Pixels \\ (Cross-Regional)}} & \multicolumn{1}{c}{\makecell{FLAIR \\ (Cross-Regional)}} & \multicolumn{1}{c}{\makecell{LoveDA \\ (Cross-Style)}} & \multicolumn{1}{c}{\makecell{Potsdam\&Vaihingen \\ (Cross Spectral Band)}} & \multicolumn{1}{c}{\makecell{OpenEarthMap \\ (Cross-Continent)}} \\
\cmidrule(lr){2-2} \cmidrule(lr){3-3} \cmidrule(lr){4-4} \cmidrule(lr){5-5} \cmidrule(lr){6-6} \cmidrule(lr){7-7} \cmidrule(lr){8-8}
& mIoU \qquad mAcc & mIoU \qquad mAcc & mIoU \qquad mAcc & mIoU \qquad mAcc & mIoU \qquad mAcc & mIoU \qquad mAcc & mIoU \qquad mAcc \\
\midrule

\rowcolor{gray!10}
\multicolumn{8}{@{}l}{\itshape\textcolor[rgb]{0.5,0.5,0.5}{Spectral Remote Sensing Domain Generalized Semantic Segmentation Models}} \\ \addlinespace[0.2em]
DSTC~\cite{liu2024dual} & 43.09 \qquad 57.81 & 29.79 \qquad 40.61 & 46.27 \qquad 59.71 & 53.42 \qquad 68.05 & 47.78 \qquad 62.90 & 16.62 \qquad 32.54 & 46.49 \qquad 63.17 \\
\midrule

\rowcolor{gray!10}
\multicolumn{8}{@{}l}{\itshape\textcolor[rgb]{0.5,0.5,0.5}{Remote Sensing Foundation Models (Freeze Backbone)}} \\ \addlinespace[0.2em]
SoftCon~\cite{wang2024multi} & {--} \qquad \qquad {--} & 29.59 \qquad 40.81 & 43.24 \qquad 58.65 & 53.21 \qquad 68.27 & 41.01 \qquad 54.91 & 11.27 \qquad 24.05 & 39.63 \qquad 60.15 \\
Galileo~\cite{tseng2025galileo} & {--} \qquad \qquad {--} & 31.68 \qquad 43.36 & 38.54 \qquad 52.63 & 52.07 \qquad 67.58 & 32.45 \qquad 45.49 & 11.69 \qquad 27.12 & 35.95 \qquad 57.12 \\
SenPaMAE~\cite{prexl2024senpa} & 45.76 \qquad 59.01 & 32.28 \qquad 45.50 & 43.31 \qquad 58.94 & 55.48 \qquad 70.34 & 38.71 \qquad 53.10 & 19.60 \qquad 34.40 & 44.13 \qquad 60.56 \\
Copernicus~\cite{wang2025towards} & 42.87 \qquad 56.83 & 33.84 \qquad 42.44 & 43.56 \qquad 60.35 & 56.15 \qquad 71.84 & 43.96 \qquad 58.16 & 15.91 \qquad 30.76 & 48.37 \qquad 69.16 \\
DOFA~\cite{xiong2024neural} & 46.03 \qquad 59.73 & 44.60 \qquad 58.66 & 49.48 \qquad 67.38 & 58.25 \qquad 72.67 & 52.41 \qquad 67.54 & 38.10 \qquad 58.74 & 58.49 \qquad 73.70 \\
\midrule

\rowcolor{gray!10}
\multicolumn{8}{@{}l}{\itshape\textcolor[rgb]{0.5,0.5,0.5}{Visual Foundation Models (Freeze Backbone)}} \\ \addlinespace[0.2em]
CLIP~\cite{radford2021learning} & {--} \qquad \qquad {--} & 46.30 \qquad 61.59 & 48.10 \qquad 64.51 & 56.81 \qquad 70.24 & 51.62 \qquad 66.95 & 40.84 \qquad 64.17 & 61.93 \qquad 78.12 \\
SAM~\cite{kirillov2023segment} & {--} \qquad \qquad {--} & 44.98 \qquad 59.35 & 47.14 \qquad 64.68 & 56.65 \qquad 71.92 & 52.17 \qquad 66.82 & 40.48 \qquad 63.57 & 56.45 \qquad 73.32 \\
EVA02~\cite{fang2024eva, fang2023eva} & {--} \qquad \qquad {--} & 45.48 \qquad 59.11 & 49.33 \qquad 62.66 & 57.52 \qquad 71.59 & 53.04 \qquad 65.70 & 42.81 \qquad 65.27 & 61.83 \qquad 75.45 \\
DINOv2~\cite{oquab2023dinov2} & {--} \qquad \qquad {--} & 53.37 \qquad 69.10 & 52.48 \qquad 69.92 & 58.38 \qquad 73.03 & 54.10 \qquad 68.39 & 55.39 \qquad 77.68 & 64.37 \qquad 76.79 \\
DINOv3~\cite{simeoni2025dinov3} & {--} \qquad \qquad  {--} & 55.54 \qquad 69.18 & 54.44 \qquad 70.71 & 59.60 \qquad 73.40 & 55.75 \qquad 74.83 & 58.79 \qquad 80.27 & 65.48 \qquad 77.02 \\
\midrule

\rowcolor{gray!10}
\multicolumn{8}{@{}l}{\itshape\textcolor[rgb]{0.5,0.5,0.5}{Foundation model-based DGSS Models}} \\ \addlinespace[0.2em]
SET~\cite{yi2024learning} & 47.38 \qquad 60.61 & 55.73 \qquad 70.24 & 55.61 \qquad 70.98 & 58.84 \qquad 73.23 & 54.58 \qquad 72.38 & 55.80 \qquad 78.85 & 64.41 \qquad 78.60 \\
FADA~\cite{bi2024learning} & 53.51 \qquad 67.54 & 56.84 \qquad 73.44 & 55.35 \qquad 71.99 & 60.12 \qquad 74.11 & 55.63 \qquad 70.85 & 59.33 \qquad 80.11 & 66.08 \qquad 78.87 \\
REIN~\cite{wei2024stronger} & 48.71 \qquad 63.58 & 59.06 \qquad 73.44 & 55.27 \qquad 71.87 & 60.46 \qquad 73.76 & 56.96 \qquad 71.37 & 60.54 \qquad 81.36 & 66.76 \qquad 78.81 \\
DepthForge~\cite{chen2025stronger} & 56.61 \qquad 69.76 & 58.79 \qquad 75.08 & 54.92 \qquad 71.43 & 61.56 \qquad 74.14 & 57.50 \qquad 71.92 & 59.57 \qquad 81.51 & 66.85 \qquad 79.70 \\
\midrule

\multicolumn{8}{@{}l}{\itshape\textcolor[rgb]{0.5,0.5,0.5}{Ours}} \\ \addlinespace[0.2em]
\rowcolor{gray!10}
SpectralMoE & \bfseries 59.83 \qquad \bfseries 73.13 & \bfseries 66.19 \qquad \bfseries 77.26 & \bfseries 60.32 \qquad \bfseries 75.78 & \bfseries 63.18 \qquad \bfseries 76.52 & \bfseries 59.11 \qquad \bfseries 75.38 & \bfseries 64.99 \qquad \bfseries 85.43 & \bfseries 68.57 \qquad \bfseries 80.17 \\
\bottomrule
\end{tabular}
\vspace{-2mm}
\caption{Quantitative comparison with SOTA methods on a comprehensive suite of DGSS benchmarks. The evaluation spans seven challenging tasks across hyperspectral, multispectral, and RGB remote sensing imagery, covering various domain shifts like cross-regional, cross-sensor, and cross-style. Our method, \textbf{SpectralMoE}, consistently establishes a new SOTA, outperforming traditional DGSS methods, frozen FM baselines (both RSFMs and VFMs), and recent FM-based DGSS adapters. The best results are highlighted in \textbf{bold}.}
\label{tab:compare_dgss}
\vspace{-4mm}
\end{table*}

\section{Experiments}
\label{sec:experiment}

\subsection{Evaluation Protocols}
\noindent \textbf{Datasets and Evaluation Metrics.}
We conduct a comprehensive evaluation on a diverse suite of seven domain generalization benchmarks (Tab.~\ref{tab:compare_dgss}), spanning hyperspectral, multispectral, and RGB RS imagery. For hyperspectral imagery, we establish a \textit{cross-region} task on the WHU-OHS dataset~\cite{li2022whu} (40 locations). For multispectral imagery, we establish three benchmarks: a \textit{cross-sensor} task on Five-Billion-Pixels~\cite{tong2023enabling} (GF-2 $\rightarrow$ GF-1/Planet/Sentinel-2); and two \textit{cross-region} tasks on geographically non-overlapping splits of Five-Billion-Pixels and FLAIR~\cite{garioud2023flair}. For RGB imagery, we set up three tasks: a rural-to-urban \textit{cross-style} task on LoveDA~\cite{wang2021loveda}; a \textit{cross-spectral-band} task (Potsdam RGB $\rightarrow$ Vaihingen NIR-R-G); and a \textit{cross-continent} task on OpenEarthMap~\cite{xia2023openearthmap}. We report mean Intersection-over-Union (mIoU) and mean Accuracy (mAcc) as primary metrics.

\noindent \textbf{Implementation Details.}
All methods are implemented within the MMSegmentation codebase for fair comparison. We employ DOFA~\cite{xiong2024neural} (a RSFM) for the hyperspectral task and DINOv3~\cite{simeoni2025dinov3} (a VFM) for multispectral and RGB tasks. We adapt DINOv3 to multispectral data by interpolating its input embedding weights. Across all tasks, we extract depth features (from RGB bands) using the PromptDA~\cite{Lin_2025_CVPR} DFM and use the Mask2former~\cite{cheng2022masked} decoder. All models are trained for 20 epochs using AdamW (learning rate $1 \times 10^{-4}$, batch size 8). Inputs are resized to $512 \times 512$, and standard augmentations include multi-scale resizing, random cropping, and horizontal flipping.

\subsection{Comparison with State-of-the-Art}
\label{exp:compare_sota}

We benchmark SpectralMoE against four categories of leading methods: (1) traditional DGSS; (2) RSFM-based; (3) VFM-based; and (4) FM-specific DGSS adapters. Full results are in Tab.~\ref{tab:compare_dgss}.

\noindent \textbf{Performance on Hyperspectral Generalization.}
On the hyperspectral cross-region task, the RSFM DOFA is the strongest baseline. While other DOFA-based adapters (SET~\cite{yi2024learning}, FADA~\cite{bi2024learning}, REIN~\cite{wei2024stronger}) improve on this, DepthForge~\cite{chen2025stronger} is the strongest competitor among them. Our SpectralMoE (also DOFA-based) achieves SOTA results, improving mIoU by 3.22\% over DepthForge and 13.8\% over the baseline (see Fig.~\ref{fig:GID} for qualitative validation).

\noindent \textbf{Performance on Multispectral Generalization.}
An interesting phenomenon is observed in the multispectral tasks: after adapting the DINOv3 VFM to multispectral data by interpolating its input embedding weights, its performance as a baseline significantly surpasses that of the RSFM DOFA. On the cross-sensor, Five-Billion-Pixels (cross-region), and FLAIR (cross-region) benchmarks, the DINOv3 baseline leads the DOFA baseline by 10.94\%, 4.96\%, and 1.35\% in mIoU, respectively. We hypothesize that this is attributable to the vastly larger scale of pre-training data used for VFMs (billions of images) compared to RSFMs (millions), enabling them to learn more robust and generalizable features. Building upon this powerful baseline, our SpectralMoE further unleashes the potential of DINOv3, outperforming the best competing methods (REIN, SET, and DepthForge) by 7.13\%, 4.71\%, and 1.62\% in mIoU across the three multispectral benchmarks.

\noindent \textbf{Performance on RGB RS Generalization.}
Consistent with the findings on multispectral tasks, the DINOv3 baseline also outperforms the DOFA baseline on the three RGB generalization tasks (cross-style, cross-spectral-band, and cross-continent). With DINOv3 as the backbone, our SpectralMoE achieves the best performance across all three tasks, surpassing the strongest competitors on each task (DepthForge, REIN, and DepthForge) by 1.61\%, 4.45\%, and 1.72\% in mIoU, respectively.

\noindent \textbf{Robustness on Different VFMs.}
To verify the robustness and applicability of SpectralMoE across different VFMs, we present a comparison in Tab.~\ref{tab:diffVFM} using CLIP~\cite{radford2021learning}, SAM~\cite{kirillov2023segment}, EVA02~\cite{fang2024eva}, and DINOv2~\cite{oquab2023dinov2} as backbones. It is worth noting that for a fair comparison of fine-tuning module efficiency, the reported parameter counts include only the adapter modules, excluding the fixed decoder parameters (20.6M). The results show that all fine-tuning methods outperform the baseline of only fine-tuning the decoder. Furthermore, SpectralMoE consistently achieves superior performance over all competing methods across all four VFM backbones, demonstrating its strong applicability. Overall, all methods deliver their best performance when using DINOv2 as the backbone.

\noindent \textbf{Comparison with General-Purpose PEFT Methods.}
To further validate our approach, we benchmark SpectralMoE against several leading general-purpose Parameter-Efficient Fine-Tuning (PEFT) methods on both DINOv3 and DOFA backbones (Tab.~\ref{tab:peft_dif_backbone}). The results show that SpectralMoE consistently outperforms all competing methods, including both general-purpose adapters and other domain generalization PEFTs, highlighting the superiority of its design.

\begin{table}[!tb]
\centering
\scriptsize 
\setlength{\tabcolsep}{2.5pt} 
\renewcommand{\arraystretch}{0.9} 

\sisetup{detect-weight, mode=text}
\renewrobustcmd{\bfseries}{\fontseries{b}\selectfont}
\begin{tabular}{@{} l l S[table-format=2.2] S[table-format=2.2] S[table-format=2.2] S[table-format=2.2] S[table-format=2.2] @{}}
\toprule
\multirow{3}{*}{\makecell[l]{Dataset}} & \multirow{3}{*}{Method} & \multicolumn{1}{c}{\multirow{3}{*}{\makecell{Params \\ (M)}}} & \multicolumn{2}{c}{DINOv3 {\tiny Large}~\cite{simeoni2025dinov3}} & \multicolumn{2}{c}{DOFA {\tiny Large}~\cite{xiong2024neural}} \\
& & & \multicolumn{2}{c}{\textit{\textcolor{gray}{\tiny(VFM)}}} & \multicolumn{2}{c}{\textit{\textcolor{gray}{\tiny(RSFM)}}} \\
\cmidrule(lr){4-5} \cmidrule(lr){6-7}
& & & {{mIoU}} & {{mAcc}} & {{mIoU}} & {{mAcc}} \\
\midrule

\multirow{9}{*}{\makecell[l]{FBP \\ \tiny(CS)}}
& Freeze & 0.00 & 53.37 & 69.10 & 46.98 & 62.83 \\
& + LoRA~\cite{hu2022lora} & 0.79 & 59.64 & 74.17 & 49.49 & 62.12 \\
& + AdaptFormer~\cite{chen2022adaptformer} & 3.22 & 57.73 & 73.22 & 49.88 & 65.89 \\
& + VPT~\cite{jia2022visual} & 3.69 & 58.14 & 72.50 & 49.70 & 60.96 \\
& + SET & 6.13 & 58.02 & 72.73 & 49.28 & 61.51 \\
& + FADA & 11.65 & 58.27 & 73.72 & 53.93 & 67.32 \\
& + REIN & 2.99 & 61.79 & 76.10 & 52.72 & 64.50 \\
& + DepthForge & 2.99 & 62.16 & 74.91 & 52.76 & 65.74 \\
\rowcolor{gray!10}
& \bfseries + SpectralMoE & 5.74 & \bfseries 66.19 & \bfseries 77.26 & \bfseries 57.73 & \bfseries 70.37 \\
\addlinespace

\multirow{9}{*}{\makecell[l]{FBP \\ \tiny(CR)}}
& Freeze & 0.00 & 52.48 & 69.92 & 49.16 & 65.41 \\
& + LoRA~\cite{hu2022lora} & 0.79 & 56.15 & 73.33 & 50.60 & 65.09 \\
& + AdaptFormer~\cite{chen2022adaptformer} & 3.22 & 55.82 & 74.89 & 50.32 & 64.99 \\
& + VPT~\cite{jia2022visual} & 3.69 & 56.04 & 73.75 & 49.43 & 68.60 \\
& + SET & 6.13 & 56.06 & 72.53 & 50.11 & 64.64 \\
& + FADA & 11.65 & 56.25 & 72.99 & 54.20 & 70.83 \\
& + REIN & 2.99 & 57.09 & 74.00 & 53.74 & 70.88 \\
& + DepthForge & 2.99 & 57.54 & 72.14 & 53.95 & 69.33 \\
\rowcolor{gray!10}
& \bfseries + SpectralMoE & 5.74 & \bfseries 60.32 & \bfseries 75.78 & \bfseries 55.41 & \bfseries 72.71 \\
\addlinespace

\multirow{9}{*}{\makecell[l]{FLAIR \\ \tiny(CR)}}
& Freeze & 0.00 & 58.38 & 73.03 & 58.25 & 72.67 \\
& + LoRA~\cite{hu2022lora} & 0.79 & 61.12 & 75.14 & 60.31 & 74.03 \\
& + AdaptFormer~\cite{chen2022adaptformer} & 3.22 & 60.19 & 72.76 & 59.68 & 74.19 \\
& + VPT~\cite{jia2022visual} & 3.69 & 60.16 & 74.11 & 59.47 & 73.36 \\
& + SET & 6.13 & 60.32 & 72.35 & 59.93 & 74.79 \\
& + FADA & 11.65 & 61.61 & 76.31 & 60.15 & 74.54 \\
& + REIN & 2.99 & 62.79 & 76.49 & 60.94 & 74.90 \\
& + DepthForge & 2.99 & 62.47 & 76.68 & 60.52 & 74.57 \\
\rowcolor{gray!10}
& \bfseries + SpectralMoE & 5.74 & \bfseries 63.18 & \bfseries 76.52 & \bfseries 61.50 & \bfseries 75.43 \\

\bottomrule
\end{tabular}
\vspace{-2mm}
\caption{
    \textbf{Comparison of PEFT methods on VFM and RSFM backbones.} Abbreviations: Five-Billion-Pixels (FBP), Cross-Sensor (CS), Cross-Regional (CR). Best results are in \textbf{bold}.
}
\label{tab:peft_dif_backbone}
\vspace{-4mm}
\end{table}

\subsection{Ablation Studies}
\label{exp:ablation_study}

\noindent \textbf{The MoE Mechanism is Crucial for Fine-Grained Feature Refinement.}
We validate the critical role of the MoE mechanism for spatially-adaptive refinement in Tab.~\ref{tab:ablation} (DINOv3 backbone). Degenerating the module to a single expert (w/o MoE), which forces a global, homogeneous adjustment strategy, causes a significant performance drop, with mIoU decreasing by 2.78\%, 0.95\%, and 2.36\% on the three multispectral benchmarks. 
This confirms that local precise refinement is essential for mitigating inter-class confusion in spatially heterogeneous spectral RS imagery.

We also investigate the dual-gated design. Forcing visual and depth features to share a single gating network (w/o Dual Gating) proved inferior to our full model, dropping mIoU by 2.75\%, 1.35\%, and 1.23\% across the tasks. This confirms our hypothesis that dedicated, modality-specific routing pathways are fundamental to preventing "modal interference" from disparate feature distributions and achieving precise expert assignment.

\noindent \textbf{Structure-Aware Feature Fusion via Cross-Attention Further Boosts Generalization.}
We next ablate the fusion components (Tab.~\ref{tab:ablation}). Removing the depth input (w/o Depth Feature) causes a substantial mIoU drop of 2.67\%, 1.31\%, and 0.11\%. This confirms our motivation that domain-robust structural information from depth is a critical supplement to spectral data. Similarly, replacing our cross-attention with simple addition (w/o Cross-Attention) is suboptimal, dropping mIoU by 2.71\%, 1.97\%, and 1.88\%. 
This demonstrates that unlike simple fusion, our cross-attention mechanism allows visual features to adaptively "query" and aggregate the most valuable structural information from the depth modality, enabling superior feature enhancement.

\begin{figure}[t]
    \centering
    \hspace*{-0.07\linewidth} 
	\includegraphics[ width=0.9\linewidth]{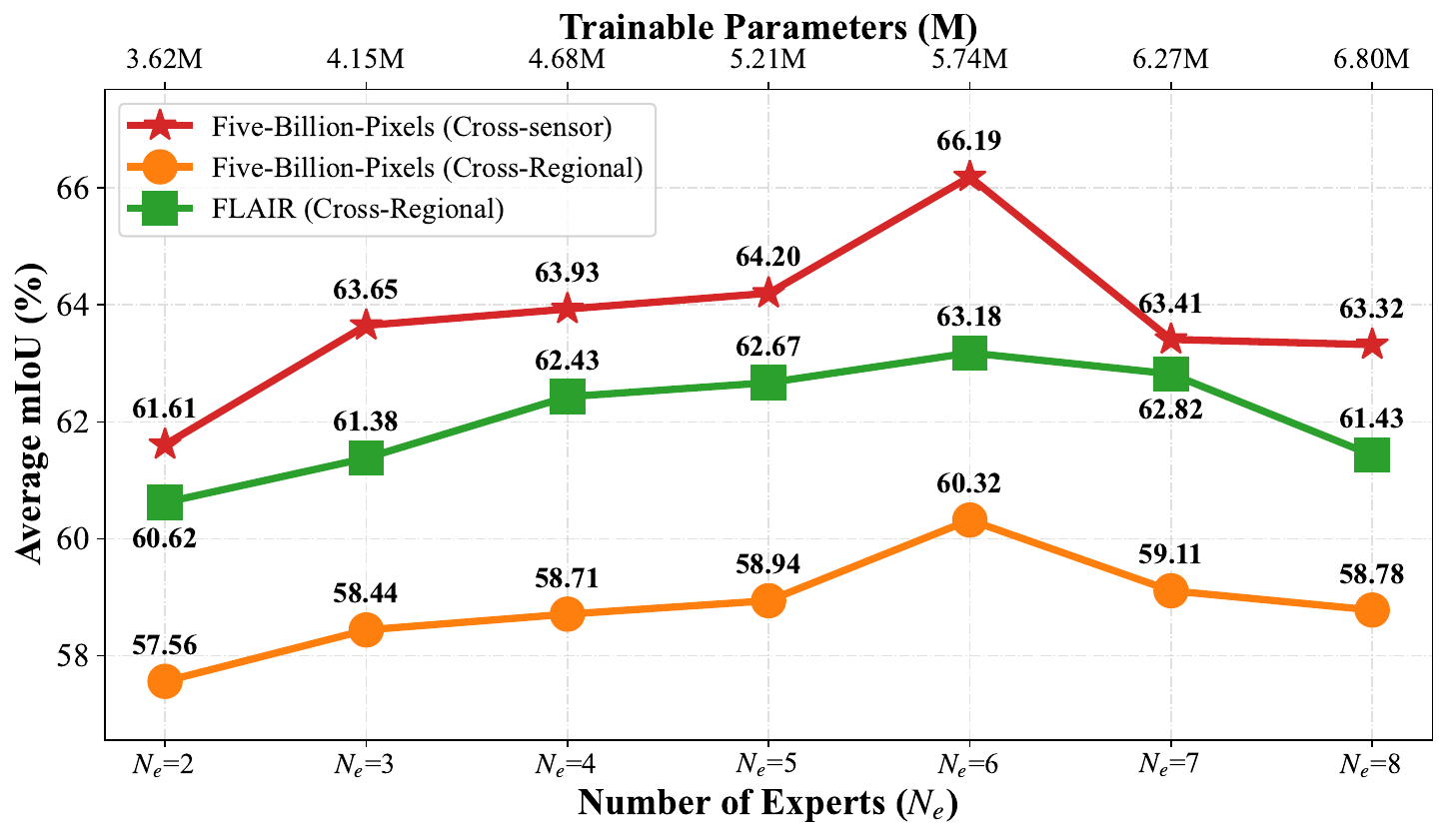}
    \vspace{-5pt}
	\caption{Ablation study on the number of experts ($N_e$).}
    \label{fig:num_experts}
    \vspace{-4mm}
\end{figure}

\noindent \textbf{Impact of the Number of Experts.}
We investigate the impact of the number of experts ($N_e$) in Fig.~\ref{fig:num_experts}. Performance is non-monotonic, peaking at $N_e=6$. Beyond this, performance degrades, which we attribute to functional redundancy and insufficient training as data is spread too thinly. Considering the trade-off between performance and linear parameter growth, $N_e=6$ is adopted as the optimal configuration, balancing model capacity and efficiency. 

\begin{table}[!tb]
\centering
\scriptsize 
\setlength{\tabcolsep}{2.5pt} 

\sisetup{detect-weight, mode=text}
\renewrobustcmd{\bfseries}{\fontseries{b}\selectfont}

\begin{tabular}{@{} l l S[table-format=2.2] S[table-format=2.2] S[table-format=2.2] @{}}
\toprule
\multirow{2}{*}{Backbone} & \multirow{2}{*}{Configuration} & \multicolumn{3}{c}{mIoU (\%)} \\
\cmidrule(lr){3-5}
& & {\makecell{\tiny Five-Billion-Pixels \\ \tiny (Cross-Sensor)}} & {\makecell{\tiny Five-Billion-Pixels \\ \tiny (Cross-Regional)}} & {\makecell{\tiny FLAIR \\ \tiny (Cross-Regional)}} \\
\midrule

\multirow{5}{*}{\makecell[l]{\textit{\textcolor{gray}{VFM}} \\ DINOv3 \\ \tiny(Large)~\cite{simeoni2025dinov3}}}
& \makecell[l]{w/o MoE} & 63.41 & 59.37 & 60.82 \\
& \makecell[l]{w/o Dual Gating} & 63.44 & 58.97 & 61.95 \\
& \makecell[l]{w/o Depth Feature} & 63.52 & 59.01 & 63.07 \\
& \makecell[l]{w/o Cross-Attention} & 63.48 & 58.35 & 61.30 \\
\rowcolor{gray!10}
& \makecell[l]{\bfseries SpectralMoE (ours)} & \bfseries 66.19 & \bfseries 60.32 & \bfseries 63.18 \\
\cmidrule(lr){2-5}

\multirow{5}{*}{\makecell[l]{\textit{\textcolor{gray}{RSFM}} \\ DOFA \\ \tiny(Large)~\cite{xiong2024neural}}}
& \makecell[l]{w/o MoE} & 55.69 & 54.58 & 59.94 \\
& \makecell[l]{w/o Dual Gating} & 53.66 & 54.10 & 60.18 \\
& \makecell[l]{w/o Depth Feature} & 54.71 & 54.44 & 61.31 \\
& \makecell[l]{w/o Cross-Attention} & 54.20 & 53.17 & 60.10 \\
\rowcolor{gray!10}
& \makecell[l]{\bfseries SpectralMoE (ours)} & \bfseries 57.73 & \bfseries 55.41 & \bfseries 61.50 \\

\bottomrule
\end{tabular}
\vspace{-2mm}
\caption{
    \textbf{Ablation study} on the components of our proposed SpectralMoE. 
    The final configuration is highlighted in \textbf{bold}.
}
\label{tab:ablation}
\vspace{-4mm}
\end{table}
\section{Conclusion}
\label{sec:conclusion}

In this paper, we addressed the critical challenge of DGSS for spectral RS, where performance is limited by spectral shifts and spatial heterogeneity. 
We identified that existing fine-tuning methods, which rely on global, homogeneous adjustments, are a key bottleneck that leads to inter-class confusion.
We proposed SpectralMoE, a novel framework that shifts the paradigm to spatially-adaptive, conditional computation. 
Our method employs a dual-gated MoE to perform local precise refinement on both visual features and depth-derived structural priors. 
A subsequent cross-attention mechanism judiciously fuses these structural cues, effectively mitigating semantic ambiguity.
Extensive experiments on a diverse suite of seven DGSS benchmarks, spanning hyperspectral, multispectral, and RGB imagery, confirmed that SpectralMoE achieves new state-of-the-art performance. 
Comprehensive ablation studies validated the efficacy of each core design choice, demonstrating that the dual-gated MoE, the introduction of depth feature priors, and the cross-attention fusion are all essential components contributing to the model's robust generalization.

\section{Acknowledgements}
This work was supported by the National Natural Science Foundation of China (NSFC) under Grant 62406331.
{
    \small
    \bibliographystyle{ieeenat_fullname}
    \bibliography{main}
}


\end{document}